%% file: acl_latex.tex
\pdfoutput=1
\documentclass[11pt]{article}

\usepackage{acl}

\usepackage{times}
\usepackage{latexsym}

\usepackage[T1]{fontenc}
\usepackage[utf8]{inputenc}

\usepackage{microtype}

\usepackage{inconsolata}

\usepackage{graphicx}
\usepackage{amsmath,amssymb,amsthm}
\usepackage{enumitem}

\usepackage{booktabs}
\usepackage{array}
\usepackage{tabularx}
\usepackage{placeins}

\title{MOOSE-Copilot: A Web-Based Interactive Assistant for Unified Exploratory and Fine-Grained Scientific Hypothesis Discovery}

\author{Hongran An \\
  Central Conservatory of Music \\
  \texttt{hongran761@gmail.com} \\\And
  Zonglin Yang$^\dagger$ \\
  Nanyang Technological University \\
  \texttt{zonglin001@ntu.edu.sg} \\}

\begin{document}
 \maketitle
\begin{abstract}
Large language models (LLMs) show remarkable potential in scientific hypothesis discovery. However, existing approaches face two critical limitations: they treat divergent exploratory search and convergent fine-grained refinement as isolated tasks, and they operate autonomously with little to no human guidance. We present MOOSE-Copilot, the first unified framework to bridge this abstraction gap through a formalized human–AI interaction (HAII) protocol. Our system empowers scientists to steer the generative process via three explicit signals: initial blueprints, inter-stage routing, and intra-stage feedback. Using an oracle-simulated evaluation in which an LLM provides idealized expert signals, we show that injecting these structured signals significantly outperforms purely autonomous baselines, characterizing the gains achievable under high-quality guidance. Furthermore, we build a web-based interface that turns the framework into a no-code workflow: researchers pose a question, watch the hypothesis search unfold as an interactive tree, and steer it by selecting hypotheses, routing between stages, and injecting feedback—no command-line agents required. This makes end-to-end hypothesis discovery directly accessible to interdisciplinary researchers.
\thanks{Website:  \url{https://moosedemo.com}}
\thanks{Video: \url{www.youtube.com/watch?v=_GSa-42ArIA}}
\let\thefootnote\relax\footnote{$^\dagger$Corresponding author.}
\end{abstract}

\section{Introduction}

Recently, large language models (LLMs) have demonstrated remarkable potential in assisting various stages of scientific research, including hypothesis generation, experiment design and execution, manuscript writing, and peer review~\citep{survey}. Among these, scientific hypothesis discovery lies at the heart of the research process, as it plays a critical role in shaping research direction and influencing the potential significance and impact of the resulting work.

We conceptualize the landscape of automated scientific hypothesis discovery as encompassing two distinct paradigms: exploratory and fine-grained discovery. Exploratory discovery focuses on broadly diverging to generate diverse, high-level research directions from a given background context. However, this often yields rough, underspecified ideas. Most existing agentic systems for scientific discovery operate within this exploratory regime~\citep{moose,DBLP:journals/corr/abs-2404-04326,li2024learning,DBLP:journals/corr/abs-2404-18400,hu2024nova,pu2024ideasynth}. 
Among these, MOOSE-Chem~\citep{msc} is the first to formalize exploratory discovery as an explicit search problem.

In contrast, fine-grained discovery—recently introduced by MOOSE-Chem2~\citep{msc2}—operates convergently. It builds upon an initial conceptual blueprint, leveraging LLMs to optimize plausibility and coherence while systematically enriching the idea with methodological and experimental details. This process ultimately translates an abstract inspiration into a concrete, actionable research hypothesis.

We observe that these two tasks correspond to two complementary stages of the scientific discovery process: exploration and exploitation. Exploratory discovery enables the broad search for promising directions, while fine-grained discovery deepens and refines those directions into actionable research hypotheses.

Despite their complementarity, two key gaps remain. First, it is entirely unexplored how these two stages might be synergistically combined within a unified framework. Unifying them is inherently challenging due to the abstraction gap between divergent exploratory search and convergent fine-grained refinement, yet this integration is an essential step toward a unified framework of automated scientific discovery. Second, current approaches to both tasks typically rely on standalone LLM workflows or agentic methods, offering little to no support for human input or control. Consequently, a major bottleneck persists: we lack a cohesive system that successfully bridges the exploratory and fine-grained phases while simultaneously empowering scientists to actively steer, course-correct, and validate this end-to-end automated process.

To overcome this bottleneck, we introduce a principled human–AI interaction (HAII) paradigm that formally bridges exploratory and fine-grained hypothesis discovery. We instantiate this paradigm through MOOSE-Copilot, a unified framework that keeps the human scientist in control across the two stages of exploration and exploitation. Rather than treating human-in-the-loop as a mere feature, we define a structured interaction protocol where scientists inject initial blueprints, make inter-stage routing decisions (when to move between the two stages), and provide intra-stage feedback (how to steer within a stage). 
This synergy mitigates the search space explosion and resolves the ambiguity of when to diverge versus when to refine, transforming an intractable automated process into a manageable, human-guided workflow.

To validate this unified framework, we systematically evaluate the impact of the core guiding signals—initial blueprints, inter-stage routing, and intra-stage feedback—on the quality of the final hypotheses. Using an oracle-simulated evaluation in which an LLM provides idealized expert signals, our quantitative analyses show that structured intervention significantly outperforms purely autonomous baselines, offering insights into how scientists can most effectively guide the system through these signals.

Furthermore, a theoretical interaction protocol is only impactful if it is accessible to its target audience. Recognizing that many domain experts—particularly in fields like chemistry and biology—are hindered by the barrier of command-line agentic tools, we operationalize our HAII paradigm through a comprehensive web-based graphical user interface (GUI).

Designed for usability and transparency, \textit{MOOSE-Copilot} visualizes the hypothesis search trajectory as an interactive tree. Through this tree, researchers trace the evolution of generated ideas and steer the search in either direction—drilling a promising branch down into fine-grained exploitation, or pivoting a refined hypothesis back to explore new directions—while injecting domain knowledge wherever needed, grounding the automated discovery process in human expertise.

In conclusion, this paper makes the following four contributions:

\begin{itemize}
    \item We propose the first work that unifies exploratory search and fine-grained exploitation in scientific hypothesis discovery.
    \item We formalize a principled human–AI interaction protocol that lets scientists steer the automated process through initial blueprints, inter-stage routing, and intra-stage feedback.
    \item We systematically evaluate the impact of these guiding signals under an oracle-simulated setting (idealized interventions standing in for expert input), showing that structured guidance significantly outperforms purely autonomous baselines.
    \item We operationalize this paradigm into an accessible web-based system featuring interactive tree visualization to ensure transparency and seamless human-in-the-loop control.
\end{itemize}

\section{Methodology}

\subsection{Preliminaries}

\subsubsection{Exploratory Hypothesis Discovery}
\label{sec:exploratory}
Generating a scientific hypothesis $h$ from a research background $b$ is often intractable. MOOSE-Chem~\citep{msc} addresses this through a fundamental assumption: that a majority of hypotheses can be composed from a research background $b$ and a set of inspirations $i_1, \ldots, i_k$ retrieved from the literature,
\begin{equation}
    h = f(b, i_1, \ldots, i_k),
    \label{eq:msc_assumption}
\end{equation}
where each inspiration $i \in I$ is a piece of knowledge not previously known to be related to $b$, and $I$ denotes the literature corpus serving as the inspiration space. Under this assumption, the intractable distribution $P(h \mid b)$ can be decomposed into a structured sequence of tractable sub-operations, transforming exploratory discovery into a search problem (with $h_0 = \emptyset$):
\begin{equation}
\small
P(h \mid b) \approx \prod_{j=1}^{k} P(i_j \mid b, h_{j-1}, I) \cdot P(h_j \mid b, h_{j-1}, i_j),
\label{eq:msc}
\end{equation}
This decomposition naturally defines a Markov Decision Process, in which the current hypothesis $h_{j-1}$ and the selected inspiration $i_j$ form a state--action pair: $P(i_j \mid b, h_{j-1}, I)$ is the policy that retrieves the next inspiration, and $P(h_j \mid b, h_{j-1}, i_j)$ is the state transition that composes it into an updated hypothesis,
\begin{equation}
b \xrightarrow{i_1} h_1 \xrightarrow{i_2} h_2 \xrightarrow{\cdots} h_{k-1} \xrightarrow{i_k} h_k = h.
\label{eq:markov_chain}
\end{equation}
In practice, an LLM-based agent realizes this process iteratively: at each step it retrieves several candidate inspirations $i_j \in I$ from the current hypothesis, and each retrieved inspiration is composed into a new hypothesis. A single step therefore branches into multiple successors, and repeating it across steps expands the hypotheses into a tree, where each node is an intermediate hypothesis and each root-to-node path is a distinct sequence of inspiration-driven updates. To keep this growth manageable, the agent applies beam search, using a ranking function $R(h)$ to retain only the top-scoring hypotheses at each step. We adopt MOOSE-Chem as the exploratory discovery stage of our system; accordingly, the MOOSE-Copilot interface exposes the resulting hypothesis tree and ranking for navigation (Section~\ref{sec:web_inference}).

\subsubsection{Fine-Grained Hypothesis Discovery}
\label{sec:fine_grained}
\input{Figures/method}
MOOSE-Chem2~\citep{msc2} introduced the task of fine-grained hypothesis discovery, which refines a coarse-grained hypothesis into a fully specified one to maximize its scientific plausibility and quality. The exploratory stage (Section~\ref{sec:exploratory}) produces a hypothesis $h_j$ that is, in principle, agnostic to granularity; in practice, without a dedicated specification step it typically remains coarse-grained, and we therefore treat it as the skeleton $h_c$ passed to the fine-grained stage. Concretely, given such a skeleton $h_c$, a fine-grained hypothesis is defined as
\begin{equation}
    h_f = \{h_c, d_1, \ldots, d_m\},
    \label{eq:fine_grained_prelim}
\end{equation}
where each edit $d_i$ adds, instantiates, or removes a detail relative to $h_c$, drawn from a specification space $D = \mathcal{D}(b, h_c)$ that is implicitly induced by the background $b$ and the skeleton $h_c$. Refinement is thus framed as selecting a coherent subset of edits from $D$:
\begin{equation}
    P(h_f \mid b, h_c) = P(\{d_1, \ldots, d_m\} \mid b, h_c, D).
    \label{eq:fine_grained_prelim2}
\end{equation}
Because $D$ is implicit and combinatorially large, this is cast as an optimization problem in which an LLM, acting as both the proposer of candidate edits and the evaluator that judges whether each edit improves the hypothesis, searches for a locally optimal $h_f$ whose internal reward approaches the global optimum.

To make this search tractable over a noisy reward landscape, MOOSE-Chem2 organizes the edits in $D$ into a hierarchy of abstraction levels and searches them \textit{level by level}, fixing higher-level choices before descending to finer ones. The hierarchy is domain-agnostic, ranging from high-level conceptual choices down to low-level operational specifics; in chemistry, for example, it spans from the reaction mechanism and the class of reagent required, down to the exact compounds, concentrations, and temperatures. This hierarchical ordering yields two benefits. First, each level searches only its own reduced candidate subset rather than the full space $D$, shrinking the per-step branching factor. Second, it smooths the reward landscape: the reward of a high-level choice can be viewed as an aggregate over the many fine-grained variants beneath it, so searching at higher levels first averages out the high-frequency irregularities of the detail space—a low-pass filtering effect that reduces premature convergence to poor local optima. As the only existing work on fine-grained hypothesis discovery, we adopt MOOSE-Chem2 for the fine-grained refinement stage of our system.

\subsection{Unified Hypothesis Discovery Framework}
\label{sec:unified_framework}
Existing research on hypothesis discovery falls into two complementary tasks: exploratory and fine-grained discovery. These correspond to the classical notions of \textit{exploration} and \textit{exploitation}—broadly searching for promising research directions, and deepening a chosen direction into a fully specified hypothesis. We unify them into a single framework, illustrated in Figure~\ref{fig:method}.

The \textit{exploratory stage} (left), implemented with MOOSE-Chem (Section~\ref{sec:exploratory}), searches the inspiration space $I$: at each step it selects an inspiration $i_j$ (the title and abstract of a paper) and updates the working hypothesis, expanding a tree of diverse coarse-grained research directions (skeletons $h_c$). The \textit{fine-grained stage} (right), implemented with MOOSE-Chem2 (Section~\ref{sec:fine_grained}), takes a skeleton $h_c$ as input and refines it into a fully specified hypothesis $h_f$ through hierarchical search over its induced specification space. Together, the two stages balance exploration and exploitation, transforming an initial research question into a fully specified, actionable hypothesis.

Bridging the two stages, however, requires deciding where to begin, when to switch between them, and how to steer each stage—decisions that are difficult to automate but natural for a domain expert. We therefore identify three guiding signals through which a human can steer the unified process, formalized in Section~\ref{sec:protocol}:
\begin{enumerate}
    \item \textbf{Initial blueprints} that constrain the starting region of the exploration--exploitation process.
    \item \textbf{Inter-stage routing}: when to transition between exploration and exploitation, and on which intermediate hypothesis.
    \item \textbf{Intra-stage feedback}: in which direction to continue refining within the current stage.
\end{enumerate}

\subsection{Formalizing the HAII Protocol}
\label{sec:protocol}
A complete hypothesis requires both exploratory search for a promising skeleton and fine-grained exploitation of its details. Carrying out both without human guidance, however, is computationally prohibitive. A fully specified hypothesis $h_f = \{h_c, d_1, \ldots, d_m\}$ is determined by two coupled searches: a search over the inspiration space $I$ that composes a skeleton $h_c = f(b, i_1, \ldots, i_k)$ (Section~\ref{sec:exploratory}), and a search over the specification space $D = \mathcal{D}(b, h_c)$ that selects the edits $\{d_1, \ldots, d_m\}$ refining $h_c$ into $h_f$ (Section~\ref{sec:fine_grained}). The difficulty is twofold. First, jointly searching $I$ and $D$ induces a combinatorial explosion of inspiration-driven branches and microscopic edits. Second, and more fundamentally, the specification space is not fixed but induced by the skeleton, $D = \mathcal{D}(b, h_c)$: any change to $h_c$ re-induces a different $D$ and invalidates the edits accumulated under the previous skeleton. The two searches are therefore entangled rather than separable, which is what makes an autonomous joint search intractable.

To overcome this, we formalize MOOSE-Copilot as a human-intervened search process in which expert interventions act as routing operators that prune and redirect the search trajectory. We collect these interventions into a set of three guiding signals $\mathcal{F} = \{f_{init}, f_{route}, f_{dir}\}$:
\begin{itemize}
    \item $f_{init}$ (\textbf{Initial Blueprint}): root-node constraints injected to prune the initial region of the inspiration search, biasing which skeletons $h_c$ are reachable from $b$.
    \item $f_{route}$ (\textbf{Inter-stage Routing}): a transition operator that selects a skeleton $h_c^*$ from the exploratory search and commits it as the entry point of fine-grained exploitation. The operator is symmetric: a refined hypothesis may also be elevated back to seed renewed exploration, in which case its skeleton is re-opened and the previously accumulated edits, being tied to the old skeleton, no longer directly apply once $D = \mathcal{D}(b, h_c)$ is re-induced under the new skeleton, though they may still inform the renewed search.
    \item $f_{dir}$ (\textbf{Intra-stage Feedback}): a feedback-conditioned regeneration step available within either stage. Given the current hypothesis $h_{prev}$ and a user critique, it augments the context, $b' = b \cup \{h_{prev}, \text{feedback}\}$, and re-generates the next hypothesis conditioned on $b'$, steering the output toward the direction indicated by the critique.
\end{itemize}
With these three signals, the entangled joint search no longer needs to be carried out at once. Instead, the expert first explores $I$ to obtain a skeleton, commits one via $f_{route}$, and only then refines its details over $D$—turning a single intractable search into two tractable single-stage searches, each conditioned on the guiding signals:
\begin{equation}
\small
\begin{aligned}
P(h_f \mid b, \mathcal{F}) \approx \;& \underbrace{P(h_c \mid b, I, f_{init}, f_{dir})}_{\text{exploration over } I} \\
&\cdot \; \underbrace{P\big(\{d_1, \ldots, d_m\} \mid b, h_c^*, D, f_{dir}\big)}_{\text{exploitation over } D}.
\end{aligned}
\label{eq:protocol}
\end{equation}
The three signals enter the factorization at different points. $f_{init}$ and $f_{dir}$ condition the two generative terms, shaping how each stage generates. $f_{route}$, by contrast, acts between the two terms. In the forward direction it selects one skeleton $h_c^*$ from those produced by the exploratory search and commits it as the entry point of exploitation, thereby inducing the specification space $D = \mathcal{D}(b, h_c^*)$ over which refinement proceeds. It can also operate in reverse, elevating a refined hypothesis back into the exploratory stage to seek a new skeleton; we return to the role of this reverse transition below.

The two terms correspond directly to the two stages defined earlier. The first term instantiates the exploratory decomposition of Equation~\ref{eq:msc}: a skeleton $h_c$ is composed from inspirations retrieved over $I$, bounded by $f_{init}$ and redirected by $f_{dir}$. The second term instantiates the fine-grained refinement of Equation~\ref{eq:fine_grained_prelim2}, now conditioned on $f_{dir}$: edits selected from the induced space $D$ refine $h_c^*$ into $h_f$.

Equation~\ref{eq:protocol} describes a single explore-then-exploit pass, assuming $f_{route}$ commits a suitable skeleton on the first attempt. In practice this is rarely optimal at once, as refinement over $D$ may reveal that the committed skeleton is itself limiting. This is why $f_{route}$ is bidirectional: a refined hypothesis can be elevated back to reopen exploration for a better skeleton, after which exploitation resumes over the newly induced $D$. The overall process therefore alternates between the two stages, progressively improving both the skeleton and its details rather than resolving them in a single pass.

\input{Figures/input_interface}
\section{MOOSE-Copilot Web Interface}
\label{sec:web_inference}

\input{Figures/tree_view}

\input{Figures/rank_review}

\input{Figures/feedback}
\input{Tables/experiments}

% Figure~\ref{fig:input_interface} shows the input interface of MOOSE-Copilot.
% Users can enter their LLM API credentials, specify a research question, optionally provide a literature survey, and upload a custom inspiration knowledge corpus to guide the exploratory search.

% Figure~\ref{fig:tree_view} illustrates the tree view of the generated hypotheses.
% Each node represents a hypothesis generated at a particular step, forming an interpretable search trace that visualizes the evolution of ideas through iterative inspiration-driven updates.

% Figure~\ref{fig:rank_review} presents the rank view of generated hypotheses.
% This view lists the ranked hypotheses along with their averaged self-evaluation scores produced by the LLM, allowing users to assess and compare hypothesis quality.

% Figure~\ref{fig:feedback} displays the feedback interface.
% After selecting a hypothesis, users can optionally provide feedback and choose the next step: continuing exploration using MOOSE-Chem~(MOOSE1) or transitioning to fine-grained exploitation via MOOSE-Chem2~(MOOSE2).

The interface operationalizes the three guiding signals of Section~\ref{sec:protocol}. Figure~\ref{fig:input_interface} shows the input interface: users enter their LLM API credentials, specify a research question, optionally provide a literature survey, and upload a custom inspiration corpus to guide the exploratory search---through which the initial blueprint $f_{init}$ is injected.
Figure~\ref{fig:tree_view} illustrates the tree view of the generated hypotheses, where each node is a hypothesis generated at a particular step, forming an interpretable trace of how ideas evolve through iterative inspiration-driven updates.
Figure~\ref{fig:rank_review} presents the rank view, listing hypotheses by their averaged LLM self-evaluation scores ($R(h)$) so that users can compare hypothesis quality.
Figure~\ref{fig:feedback} displays the feedback interface. After selecting a hypothesis, the user can provide intra-stage feedback ($f_{dir}$) and make the inter-stage routing decision ($f_{route}$): continuing exploration with MOOSE-Chem or transitioning to fine-grained exploitation with MOOSE-Chem2. The routing is bidirectional---a refined hypothesis can likewise be taken back into exploration to seek a new direction.

\section{Experiments}
\label{sec:experiments}
We evaluate our system on the TOMATO-Chem2 dataset~\citep{msc2}, leveraging its detailed annotations (research questions, literature surveys, and fine-grained hypotheses) across 51 top-tier papers. Following \citet{msc2}, we compute the recall of ground-truth elements recovered by the generated hypotheses (Table~\ref{tab:result}). Our experiments isolate and assess the impact of the protocol's three guiding signals: (1) initial blueprints, (2) inter-stage routing, and (3) intra-stage feedback.

\textbf{Justification for Oracle-Simulated Evaluation.}
To measure the system's intrinsic responsiveness to these signals---free from the confounding variance of human expertise---we adopt an \textit{oracle-simulated evaluation}. Feedback is provided by an oracle LLM with access to the ground-truth hypothesis but instructed to give guiding critiques without disclosing the answer, and node selection is simulated by oracle ranking against the ground truth. This characterizes the recall achievable under idealized guidance, serving as an upper bound rather than an estimate of real-world human gains.

\textbf{Results.}
Table~\ref{tab:result} supports three observations. First, \textbf{initial blueprints} improve effectiveness, raising recall from $11.44\%$ to $15.37\%$ over the MC baseline. Second, \textbf{inter-stage routing} largely determines final performance: oracle ranking instead of self-ranking at the routing step raises recall from $12.74\%$ to $18.26\%$. 
% Third, \textbf{intra-stage feedback} drives stronger refinements, with repeated strong feedback yielding the best result ($26.96\%$). The MC2 baseline alone ($10.33\%$) underperforms the MC baseline ($11.44\%$), as a refinement procedure offers no advantage without a well-formed coarse hypothesis as input. 
Third, \textbf{intra-stage feedback} drives stronger refinements, and crucially its \emph{quality} matters more than its quantity. At matched rounds, strong feedback consistently reaches higher recall than ordinary feedback while using fewer optimization steps (e.g., at four rounds, $26.96\%$ at $90.1$ steps vs.\ $20.99\%$ at $107.4$ steps). Moreover, repeating refinement under strong feedback steadily improves recall ($23.10\% \rightarrow 26.96\%$ over one to four rounds), whereas under ordinary feedback additional rounds reduce the optimization steps but no longer improve recall ($\sim\!21\%$ throughout). This indicates that high-quality critiques, rather than simply more refinement iterations, are what convert extra search into better hypotheses.
% Finally, repeating the feedback-guided refinement reduces the optimization steps needed at each round (e.g., $166.1 \rightarrow 107.4$ across one to four rounds): once earlier rounds have fixed part of the details correctly, each subsequent round only updates the remaining components, converging in progressively fewer steps.

\section{Conclusion}
We presented MOOSE-Copilot, the first framework to bridge the abstraction gap between exploratory search and fine-grained exploitation in scientific hypothesis discovery. By formalizing a principled human–AI interaction protocol, we let scientists steer the automated process via initial blueprints, inter-stage routing, and intra-stage feedback. Under an oracle-simulated evaluation, structured guidance significantly outperforms purely autonomous baselines, characterizing the gains achievable under high-quality guidance. Furthermore, our interactive web interface lowers the technical barrier of command-line agentic tools, making AI-assisted hypothesis discovery accessible to interdisciplinary researchers beyond machine-learning experts.

\FloatBarrier

% \section*{Limitations}
% MOOSE-Copilot does not currently integrate automated experiment execution~\citep{msc3,funsearch,alphaevolve}, which would close the loop between hypothesis generation and empirical validation. The system does, however, expose a hook for incorporating external experimental results through the intra-stage feedback signal $f_{dir}$, which we leave as a direction for future work.
% It also does not yet integrate post-training methods for scientific discovery~\citep{yang2026moosestar}, which we likewise leave as future work.

\section*{Limitations}
MOOSE-Copilot currently has two main limitations.
First, it does not integrate automated experiment execution~\citep{msc3,funsearch,alphaevolve}, which would close the loop between hypothesis generation and empirical validation. The system does, however, expose a hook for incorporating external experimental results through the intra-stage feedback signal $f_{dir}$.
Second, it does not yet leverage post-training methods specifically developed for scientific hypothesis discovery~\citep{yang2026moosestar}, which could improve the generation quality of both the exploratory and fine-grained stages.
We leave both as directions for future work.

\section*{Ethics Statement}

This work aims to support scientific research by developing a system that assists in hypothesis generation through human–AI collaboration. All datasets used in this study are derived from publicly available scientific literature and do not contain personally identifiable or sensitive information. The system does not produce or promote misinformation intentionally, and user interactions remain local unless explicitly shared.

While MOOSE-Copilot has the potential to accelerate scientific discovery, we acknowledge possible misuse—for example, generating misleading or unfounded hypotheses if used without proper domain knowledge or oversight. We emphasize that the system is intended to augment, not replace, expert judgment, and we encourage responsible use by qualified researchers. Future work should further investigate the societal and epistemic implications of AI-augmented scientific workflows, including issues of bias, overreliance, and reproducibility.

We have made efforts to promote transparency, reproducibility, and accessibility by releasing our code and interface under a permissive open-source license.

\section*{Broader Impact Statement}

MOOSE-Copilot aims to democratize and accelerate scientific discovery by providing researchers with a powerful, interactive assistant for hypothesis generation. By integrating both exploratory and fine-grained reasoning processes into a single human–AI framework, the system has the potential to support more systematic, creative, and efficient research workflows across a wide range of scientific disciplines, including chemistry, biology, and materials science.

This broader accessibility may empower smaller research teams, early-career scientists, and institutions with fewer computational resources to participate more actively in hypothesis-driven research. It also opens new opportunities for interdisciplinary collaboration, as researchers can more easily explore ideas outside their immediate domain expertise with LLM-assisted support.

However, there are also potential risks. Improper use of the system—such as uncritical acceptance of AI-generated hypotheses, misuse in ideologically driven or pseudoscientific contexts, or reinforcement of biases present in training data—could negatively affect the scientific process. To mitigate this, MOOSE-Copilot is designed to keep humans meaningfully in the loop, encouraging critical reflection, transparency, and control.

Overall, we view this work as a step toward more inclusive, efficient, and assistive scientific research, while recognizing the importance of responsible use, continuous evaluation, and community oversight in its deployment and development.

% \section*{Limitations}

% \section*{Acknowledgments}

\bibliography{custom}

\appendix

\end{document}

%% file: Figures/method.tex
\begin{figure*}[htbp]
\centering
\resizebox{2\columnwidth}{!}{
\includegraphics[]{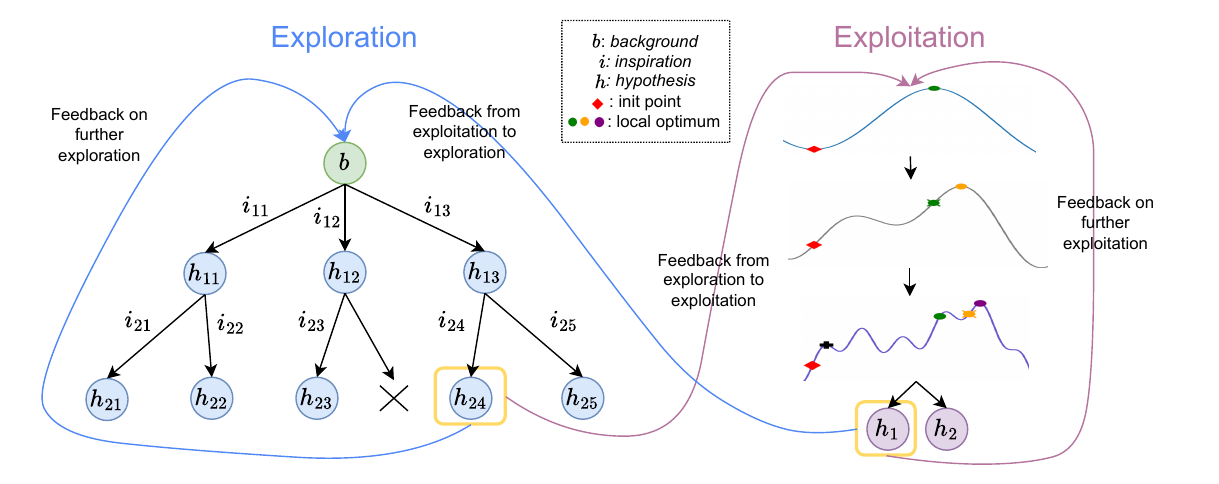}
}
\caption{Overview of MOOSE-Copilot. Yellow rectangles indicate hypotheses actively selected by the user for further exploration or exploitation. Blue and purple paths represent user-directed transitions to additional exploration or fine-grained exploitation stage, respectively. Feedback can be provided along with each additional stage selection.}
\label{fig:method}
\end{figure*}

%% file: Figures/input_interface.tex
\begin{figure}[htbp]
\centering
\resizebox{0.75\columnwidth}{!}{
\includegraphics[]{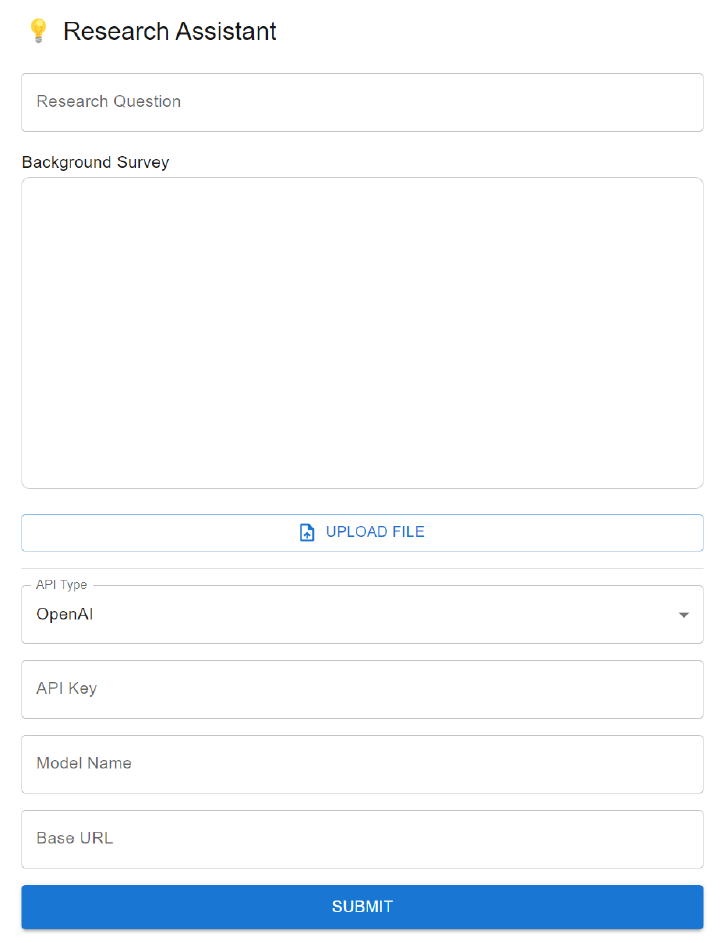}
}
\caption{Input interface of MOOSE-Copilot.}
\label{fig:input_interface}
\end{figure}

%% file: Figures/tree_view.tex
\begin{figure*}[htbp]
\centering
\resizebox{1.8\columnwidth}{!}{
\includegraphics[]{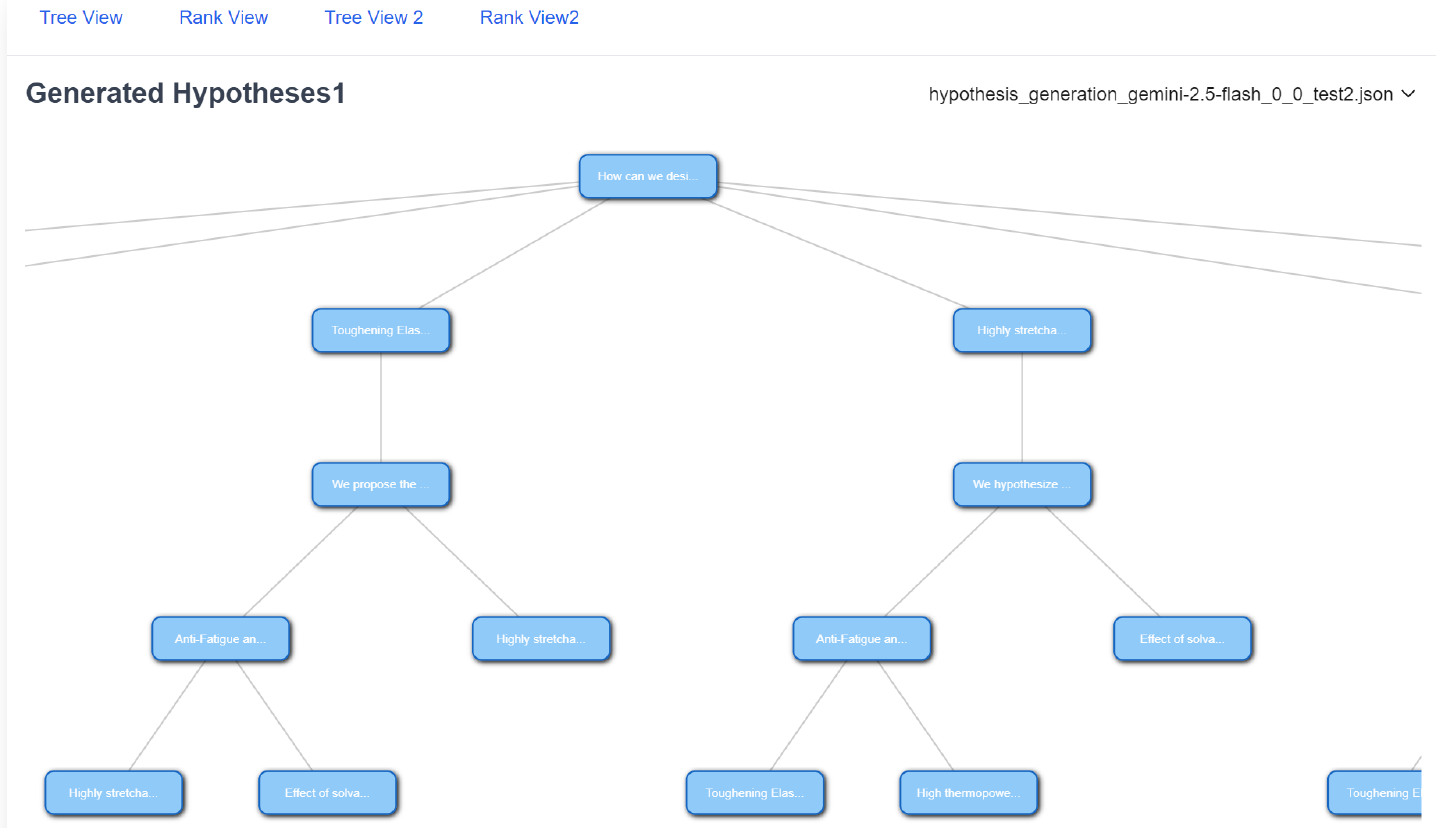}
}
\caption{Tree view of the hypothesis generation process in MOOSE-Copilot.}
\label{fig:tree_view}
\end{figure*}

%% file: Figures/rank_review.tex
\begin{figure*}[htbp]
\centering
\resizebox{2\columnwidth}{!}{
\includegraphics[]{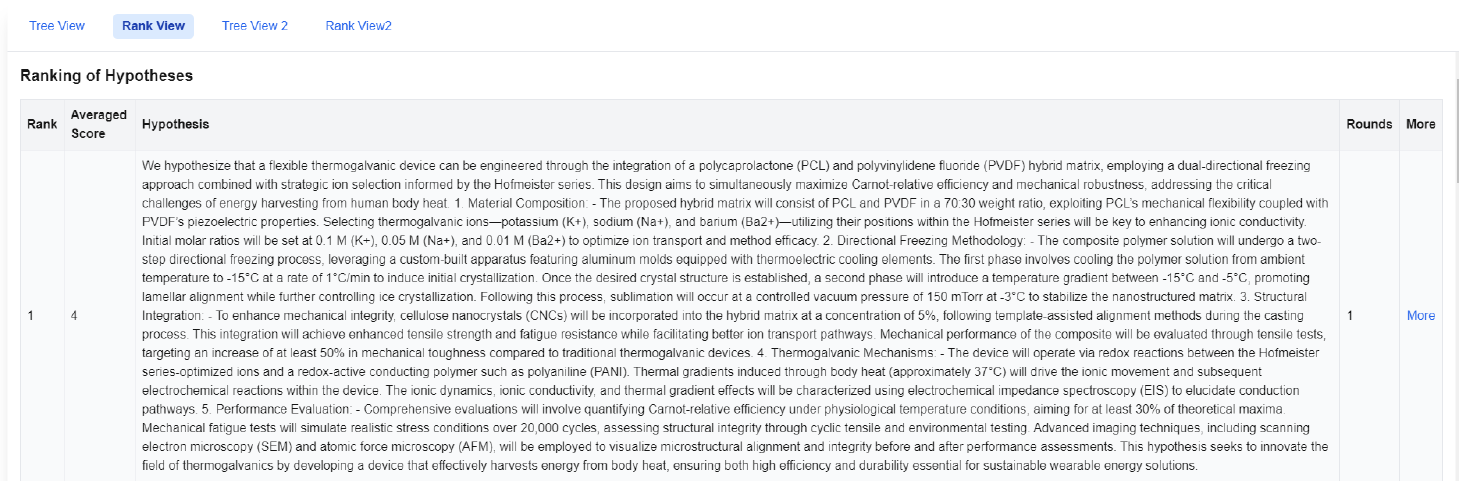}
}
\caption{Hypothesis ranking interface in MOOSE-Copilot. }
\label{fig:rank_review}
\end{figure*}

%% file: Figures/feedback.tex
\begin{figure*}[htbp]
\centering
\resizebox{1.6\columnwidth}{!}{
\includegraphics[]{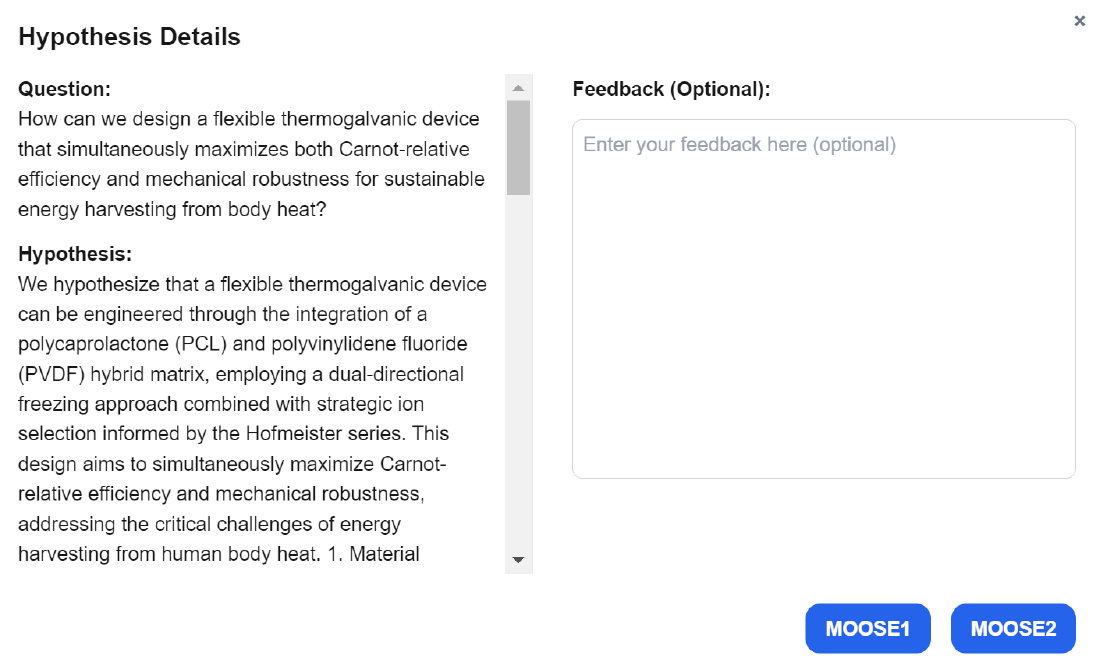}
}
\caption{Feedback interface in MOOSE-Copilot.}
\label{fig:feedback}
\end{figure*}

%% file: Tables/experiments.tex
\begin{table*}[!htbp]
\centering
\resizebox{2\columnwidth}{!}{
\begin{tabular}{l | p{8cm} |r|r}
\toprule
Method Name                                       & Description                                                                                      & Recall & \# Search Steps \\ \midrule
baseline\_MC~\citep{msc}                                      & MC                                                                                               & 11.44\%                             & -               \\
baseline\_MC2~\citep{msc2}                                     & MC2                                                                                              & 10.33\%                             & 478.6                               \\
MC\_with\_hint                                    & MC + initial blueprint                                                                             & 15.37\%                             & -               \\ \midrule
MC\_with\_soft\_feedback\_with\_hint              & MC + initial blueprint + (oracle ranking + soft feedback) + MC                                     & 16.78\%                             & -               \\
MC\_with\_feedback\_with\_hint                    & MC + initial blueprint + (oracle ranking + feedback) + MC                                          & 16.93\%                             & -               \\ \midrule
MC2\_with\_MC\_input\_self\_rank                  & MC + initial blueprint + (self-ranking) + MC2                                                      & 12.74\%                             & 321.2                               \\
MC2\_with\_MC\_input\_oracle\_rank                & MC + initial blueprint + (oracle-ranking) + MC2                                                    & 18.26\%                             & 336.6                               \\ \midrule
MC2\_with\_feedback\_oracle\_rank                 & MC + initial blueprint + (oracle-ranking) + MC2 + {[}(oracle ranking + feedback) + MC2{]}x1        & 21.98\%                             & 166.1                               \\
MC2\_with\_feedback\_x2\_oracle\_rank             & MC + initial blueprint + (oracle-ranking) + MC2 + {[}(oracle ranking + feedback) + MC2{]}x2        & 21.91\%                             & 149.9                               \\
MC2\_with\_feedback\_x3\_oracle\_rank             & MC + initial blueprint + (oracle-ranking) + MC2 + {[}(oracle ranking + feedback) + MC2{]}x3        & 22.35\%                             & 123.1                               \\
MC2\_with\_feedback\_x4\_oracle\_rank             & MC + initial blueprint + (oracle-ranking) + MC2 + {[}(oracle ranking + feedback) + MC2{]}x4        & 20.99\%                             & 107.4                               \\ \midrule
MC2\_with\_strong\_feedback\_oracle\_rank     & MC + initial blueprint + (oracle-ranking) + MC2 + {[}(oracle ranking + strong feedback) + MC2{]}x1 & 23.10\%                             & 129.4                               \\
MC2\_with\_strong\_feedback\_x2\_oracle\_rank & MC + initial blueprint + (oracle-ranking) + MC2 + {[}(oracle ranking + strong feedback) + MC2{]}x2 & 25.54\%                             & 102.1                               \\
MC2\_with\_strong\_feedback\_x3\_oracle\_rank & MC + initial blueprint + (oracle-ranking) + MC2 + {[}(oracle ranking + strong feedback) + MC2{]}x3 & 25.70\%                             & 123.1                               \\
MC2\_with\_strong\_feedback\_x4\_oracle\_rank & MC + initial blueprint + (oracle-ranking) + MC2 + {[}(oracle ranking + strong feedback) + MC2{]}x4 & 26.96\%                             & 90.1                               \\
\bottomrule
\end{tabular}}
\caption{Experimental results. ``MC'' represents MOOSE-Chem, ``MC2'' represents MOOSE-Chem2.}
\label{tab:result}
\end{table*}